\documentclass[conference]{IEEEtran}

\usepackage{cite}
\usepackage{mathtools}
\usepackage{amsmath,amssymb,amsfonts}
\usepackage{algorithmic}
\usepackage{graphicx}
\usepackage{subfigure}
\usepackage{verbatim}
\usepackage{textcomp}
\usepackage{float}
\usepackage{esvect}
\usepackage{color}

\begin{document}

\title{Camera Pose Correction in SLAM Based on Bias Values of Map Points}

\author{Zhaobing Kang, Wei Zou and Zheng Zhu}
\thanks{This work is supported by the National Natural Science Foundation of China (Grant No. 61773374), and the Project of Development in Tianjin for Scientific Research Institutes Supported By Tianjin Government (project 16PTYJGX00050).}
\thanks {Zhaobing Kang, Dengpeng Xing, and Hongxuan Ma are affiliated to Institute of Automation, Chinese Academy of Sciences, Beijing China;
        { \tt\small kangzhaobing2017@ia.ac.cn}}%
\thanks {Wei Zou is affiliated to Institute of Automation, Chinese Academy of Sciences, Beijing China and TianJin Intelligent Tech.Institute of CASIA Co. ,Ltd, Tianjin China;
        {\tt\small wei.zou@ia.ac.cn}}%

\maketitle

\begin{abstract}
Accurate camera pose estimation result is essential for visual SLAM (VSLAM). This paper presents a novel pose correction method to improve the accuracy of the VSLAM system. Firstly, the relationship between the camera pose estimation error and bias values of map points is derived based on the optimized function in VSLAM. Secondly, the bias value of the map point is calculated by a statistical method. Finally, the camera pose estimation error is compensated according to the first derived relationship. After the pose correction, procedures of the original system, such as the bundle adjustment (BA) optimization, can be executed as before. Compared with existing methods, our algorithm is compact and effective and can be easily generalized to different VSLAM systems. Additionally, the robustness to system noise of our method is better than feature selection methods, due to all original system information is preserved in our algorithm while only a subset is employed in the latter. Experimental results on benchmark datasets show that our approach leads to considerable improvements over state-of-the-art algorithms for absolute pose estimation. 
\end{abstract}
\section{INTRODUCTION}
VSLAM can estimate the camera trajectory and reconstruct environment, therefore, it is very important on many occasions, such as mobile robots navigation and augmented reality (AR). To improve the accuracy, parallelism \cite{1} and orthogonality \cite{8} of lines or planes are utilized. However, since no prior structural information is acquired when exploring new environments, they cannot be used. Unlike the above methods, the optimization-based methods \cite{3}, \cite{37} or the matrix-theory-based approach \cite{16} are not limited by the environment. But most pipelines of them are complex and difficult to combine with different SLAM systems. Multi-sensor fusion can compensate the drawbacks of each other, therefore, the IMU is widely used in VSLAM to improve the system robustness \cite{00}, \cite{34}, \cite{35}. However, accurate IMU bias estimation is difficult, and large estimation error may affect the localization performance of the SLAM system.

This paper presents a novel pose correction method, which is compact and effective and reserves all original system information. The most related work to this paper is \cite{31}, where the map points with small bias value are chosen to estimate the camera pose while the other map points are abandoned. Different from choosing a subset of map points to reduce the estimation error in \cite{31}, our method compensates the pose estimation error based on the bias values of all map points. 

Our algorithm has two advantages compared with existing methods. The first advantage is our method is compact and effective and easy to integrate into different SLAM systems. The second is the robustness of our method is better than feature selection methods, such as \cite{31}. Since only a subset of features is chosen in feature selection methods, they are more sensitive to system noise, which is demonstrated by the experimental results.

\section{Related Works}
\textbf{Structural regularity based methods: }A monocular SLAM system, which leverages structural regularity in Manhattan world and contains three optimization strategies is proposed in \cite{1}. However, to reduce the estimation error of the rotation motion, multiple orthogonal planes must be visible throughout the entire motion estimation process. Unlike only using planes in \cite{1}, the rotation motion is estimated by joint lines and planes in \cite{2}. Once the rotation is found, the translational motion can be recovered by minimizing the de-rotated reprojection error. In \cite{4}, the accuracy of BA optimization is enhanced by incorporating feature scale constraints into it. Structural constraints between nearby planes (e.g. right angle) are added in the SLAM system to further recover the drift and distortion in \cite{8}. Since the structural regularity does not exist in all environments, the application scope of this category is limited.

\textbf{Optimization-based methods and matrix-theory-based methods:} A new initialization method for the orientations of the pose graph optimization problem is proposed in \cite{3}. In this method, the orientation values are calculated by an iterative approach, and the relative orientation mismatches of the cost function are approximated by a quadratic cost function. In \cite{9}, the photometric and the depth error over all pixels are employed to reduce the estimation error in the RGB-D system. However, this method is time-consuming and difficult to achieve real-time performance. Different from using all pixel depth information in \cite{9}, a monocular camera combined with sparse depth information from LiDAR is employed in \cite{10}, and three optimization strategies are carefully designed considering both accuracy and time-consuming. Similar to \cite{9} and \cite{10}, a new approach, which utilizes dense fusion of several stereo depths in the locality establishes a locally dense and globally sparse map. Rao-Blacwellized particle filter (RBPF) method is employed in \cite{14} and \cite{29}. The difference between them is \cite{14} presents a new RBPF method while drawbacks of the RBPF are overcome in \cite{29} by scaled unscented transformation. How to obtain accurate map in the large or scale uncertain environments is studied in \cite{15,26,28}. To get a good accuracy without sacrificing speed, new matrix decompose methods are proposed in \cite{16} and \cite{22}. Different from the above methods, where all system information is employed, a good feature selection algorithm is introduced in \cite{31}. By selecting the map points which have smaller error, it can reach a balance between the error expectation and the covariance. Most pipelines of them are complex, therefore, it is difficult to integrate them into different SLAM systems.

\textbf{Methods of integration with IMU:} In \cite{6}, four cameras and an IMU are tightly fused in a Micro Aerial Vehicle (MAV). A new approach tightly combines visual measurements with IMU measurements is proposed in \cite{19}. The novelty lies in that the IMU error term is integrated with the landmark reprojection error in a fully probabilistic manner. In \cite{30}, IMU information is employed in the ORB-SLAM \cite{0} to solve the scale problem of a monocular system.
The performance of the SLAM system is easily affected by the bias estimation results and IMU noise.

The main contributions of this paper can be summarized as follows. (i) A new camera pose correction method is proposed. (ii) A bias calculating method used for the map point is integrated into our framework. Thanks to this method, our system can operate in real-time. (iii) Experimental results demonstrate that our method outperforms the state-of-the-art SLAM system.

\section{relationship between pose estimation error and map point bias}

\begin{figure} \centering 
\includegraphics[scale=0.3]{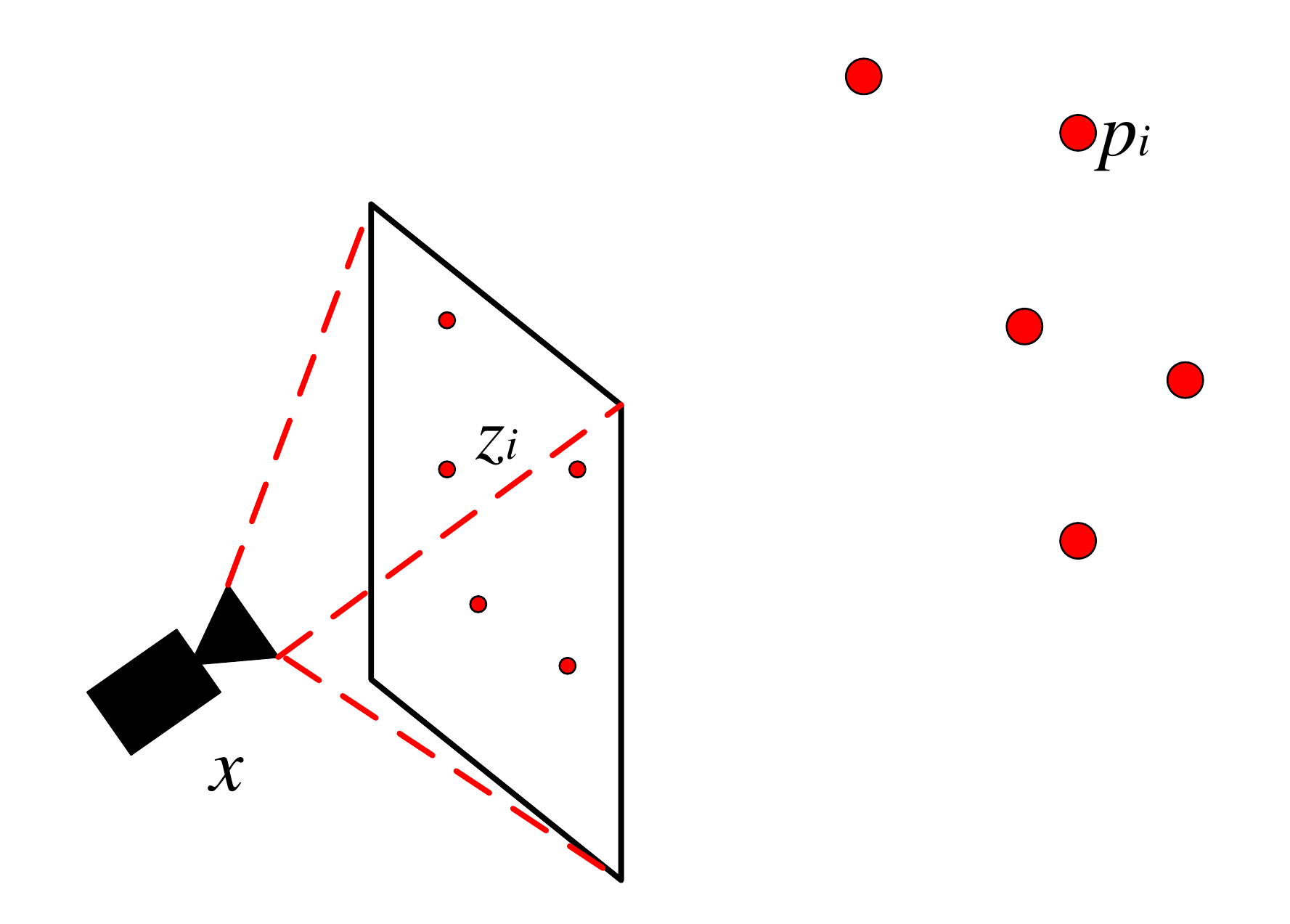} 
\caption{The commonly used SLAM system. In this figure, the map point $\emph{\textbf{p}}_i$ is projected to the image point $\emph{\textbf{z}}_i$ at the camera pose $\emph{\textbf{x}}$.}
\label{fig1} 
\end{figure}

The commonly used SLAM system is shown in Fig. 1. In this system, the main object is to estimate the camera pose $\emph{\textbf{x}}$ and the map point $\emph{\textbf{p}}_i$ by minimizing the errors between the observation values and the estimation values, which can be written as

\begin{equation}
\arg\ \min\limits_{\emph{\textbf{x}}, \emph{\textbf{p}}_i} \frac{1}{2}\sum_{i=1}^n \lVert h\left(\emph{\textbf{x}}, \emph{\textbf{p}}_i\right)-\emph{\textbf{z}}_i \rVert^2
\label{20190821eq1}
\end{equation}
where $n$ represents the number of matched image feature point $\emph{\textbf{z}}_i$ at the camera pose $\emph{\textbf{x}}$, and $h(\emph{\textbf{x}},\emph{\textbf{p}}_i)$ represents the camera projection model. To simplify the description, for the symbols whose dimension can be easily determined, the subscripts of them are omitted. This problem can be solved by the Gaussian-Newton method or the Levenburg-Marquadt (LM) method. Since $h(\emph{\textbf{x}},\emph{\textbf{p}}_i)$ is a nonlinear function, it must be linearized to fit for these optimization methods. The first-order approximation to $h(\emph{\textbf{x}},\emph{\textbf{p}}_i)$ about initial guess $\emph{\textbf{x}}^{(s)}$ can be written as
\begin{equation}
h\left(\emph{\textbf{x}}, \emph{\textbf{p}}_i\right) = h\left(\emph{\textbf{x}}^{(s)}, \emph{\textbf{p}}_i\right)+\textbf{H}_x(\emph{\textbf{x}} - \emph{\textbf{x}}^{(s)})  \label{eq2}
\end{equation}
where $\textbf{H}_x$ is the Jacobian matrix about $\emph{\textbf{x}}$. Similar to (\ref{eq2}), the first-order approximation to $h\left(\emph{\textbf{x}}^{(s)}, \emph{\textbf{p}}_i\right)$ about initial guess $\emph{\textbf{p}}_i^{(s)}$ is
\begin{equation}
h\left(\emph{\textbf{x}}^{(s)}, \emph{\textbf{p}}_i\right) = h\left(\emph{\textbf{x}}^{(s)}, \emph{\textbf{p}}_i^{(s)}\right)+\textbf{H}_{pi}(\emph{\textbf{p}}_i - \emph{\textbf{p}}_i^{(s)}) \label{eq3}
\end{equation}
where $\textbf{H}_{pi}$ is the Jacobian matrix about $\emph{\textbf{p}}_i$.

According to the Gaussian-Newton method, (\ref{20190821eq1}) and (\ref{eq2}), the pose update can be written as
\begin{equation}
\emph{\textbf{x}}^{(s+1)} = \emph{\textbf{x}}^{(s)} - \textbf{H}_x^+\left(h\left(\emph{\textbf{x}}^{(s)}, \emph{\textbf{p}}_i\right)-\emph{\textbf{z}}_i\right) \label{eq4}
\end{equation}
where $\textbf{H}_x^+$ is the pseudo-inverse of $\textbf{H}_x$. Substituting (\ref{eq3}) into (\ref{eq4}), the pose estimation error is
\begin{equation}
\epsilon_x = -\textbf{H}_x^+\left(\textbf{H}_{pi}\epsilon_{pi}+\epsilon_{zi}\right) \label{eq5}
\end{equation}
where $\epsilon_x = \emph{\textbf{x}}^{(s+1)} - \emph{\textbf{x}}^{(s)}$ represents the pose estimation error, $\epsilon_{zi} = h\left(\emph{\textbf{x}}^{(s)}, \emph{\textbf{p}}_i^{(s)}\right) - \emph{\textbf{z}}_i$ represents the observation error, and $\epsilon_{pi} = \emph{\textbf{p}}_i - \emph{\textbf{p}}_i^{(s)}$ represents the map point bias.

Suppose the image observation error subjects to the zero-mean Gaussian distribution, i.e. $\epsilon_{zi}\sim N\left(\emph{\textbf{0}}, \sum_{zi}\right)$, and the error of the map point subjects to non-zero-mean Gaussian distribution, i.e. biased distribution $\epsilon_{pi}\sim N\left(\boldsymbol{\mu}_{pi}, \sum_{pi}\right)$. This assumption is reasonable, because the bias of the map point can be introduced by the BA optimization or the image measurement error. According to this assumption, the expectation of the camera pose estimation error is
\begin{equation}
\emph{\textbf{E}}\left[\epsilon_x\right] = -\textbf{H}_x^+\textbf{H}_p\textbf{1}\boldsymbol{\mu}_p. \label{eq6}
\end{equation}
where 
\begin{equation}
\textbf{1} = \left[\begin{matrix}\begin{smallmatrix}
1 & 1 & 1 & 0 & 0 & 0 & ... & 0 & 0 & 0\\
0 & 0 & 0 & 1 & 1 & 1 & ... & 0 & 0 & 0\\
. & . & . & . & . & . & ... & . & . & .\\
. & . & . & . & . & . & ... & . & . & .\\
. & . & . & . & . & . & ... & . & . & .\\
0 & 0 & 0 & 0 & 0 & 0 & ... & 1 & 1 & 1\\
\end{smallmatrix}\end{matrix}\right]_{3n\times 3n}.
\end{equation}
It is obvious that the pose estimation error can be reduced if the bias value of the map point is known. However, getting accurate bias value is difficult due to system errors. In this paper, a bias calculating expression proposed in \cite{41} is employed, which will be introduced in the next section.

\section{Camera Pose Correction}
\subsection{Bias Calculation}

\begin{table}[htb]
\center
\caption{The symbol meanings used in the bias expression.}\label{20190820table1}
\setlength{\tabcolsep}{2mm}{\begin{tabular}{cl}
\hline  
Symbol & Meaning\\
\hline  
$\left(x_i, y_i\right)$	&the image coordinates of the point $z_i$\\
$\left(x_f, y_f\right)$   &the focus of expansion (FOE), equals to $\left(V_x/V_z, V_y/V_z\right)$\\
$z\left(x_i, y_i\right)$  &the depth of the point $z_i$\\
$\emph{\textbf{d}} $      & $ \left(d\left(x_1,y_1\right),...,d\left(x_N,y_N\right)\right)^T $\\
$\emph{\textbf{u}}$       & $ \left(p\left(x_1,y_1\right), q\left(x_1,y_1\right),...,p\left(x_N,y_N\right), q\left(x_N,y_N\right)\right)^T $\\
$\emph{\textbf{r}}_i$     &$\left(x_iy_i, -\left(1+x_i^2\right), y_i\right)^T$\\
$\emph{\textbf{s}}_i$     &$\left(1+y_i^2, -x_iy_i, -x_i\right)^T$\\
$\boldsymbol{\Omega}$                  &$\left(\omega_x, \omega_y, \omega_z\right)^T$\\
$\textbf{Q}$                       &$\left[\emph{\textbf{r}}_1,\emph{\textbf{s}}_1,...,\emph{\textbf{r}}_N,\emph{\textbf{s}}_N\right]^T$\\
$\textbf{P}$     &diagonal matrix $diag\left[ \begin{matrix}x_i-x_f\\y_i-y_f \end{matrix}\right]_{2N\times N, i=1,...,N}$\\
$\textbf{B}$     &$\left[\textbf{P} \ \textbf{Q}\right]$\\
$\emph{\textbf{z}}$     &$\left[ \begin{matrix}\emph{\textbf{d}}\\ \boldsymbol{\Omega} \end{matrix}\right]$\\
\hline
\end{tabular}}
\end{table}

The basic idea of the bias calculating method is introduced in this subsection, and more details can be available in \cite{41}. Let $\emph{\textbf{V}} = \left[v_x, v_y, v_z\right]^T$ and $\boldsymbol{\Omega} = \left[\omega_x, \omega_y, \omega_z\right]^T$ respectively represent the translation vector and the rotation vector of the camera. According to the optical flow, 3D camera motion and scene depth, the velocity fields of the image point $z_i$ can be expressed as
\begin{equation}
\begin{aligned}
p\left(x_i,y_i\right) &= \left(x_i-fx_f\right)d(x_i,y_i)+\frac{1}{f}x_iy_i\omega_x\\
&-\left(f+\frac{1}{f}x_i^2\right)\omega_y+y_i\omega_z\\
q\left(x_i,y_i\right) &= \left(y_i-fy_f\right)d(x_i,y_i)-\frac{1}{f}x_iy_i\omega_y\\
&+\left(f+\frac{1}{f}y_i^2\right)\omega_x-x_i\omega_z
\end{aligned}\label{20190820eq1}
\end{equation}
where $p\left(x_i,y_i\right)$ and $q\left(x_i,y_i\right)$ are the horizontal and vertical velocity fields, $d(x_i,y_i)=v_z/z\left(x_i,y_i\right)$ is the scaled inverse scene depth, and the meanings of the other parameters are listed in Table \ref{20190820table1}. For $N$ matched feature points in two consecutive frames, normalizing linear distances with respect to the focal length, (\ref{20190820eq1}) can be written as a matrix form
\begin{equation}
\emph{\textbf{u}}=\textbf{P}\emph{\textbf{d}} + \textbf{Q}\boldsymbol{\Omega} = \left[\textbf{P}\ \textbf{Q}\right]\left[\begin{matrix}\emph{\textbf{d}}\\ \boldsymbol{\Omega}\end{matrix}\right] \triangleq \textbf{B}\emph{\textbf{z}}
\label{20190820eq2}
\end{equation}
where the meanings of symbols are shown in Table \ref{20190820table1}. The aim is calculating $\emph{\textbf{z}}$ from $\emph{\textbf{u}}$.
For VSLAM, since the camera motions corresponding to these two frames are known, $\boldsymbol{\Omega}$ is known. (\ref{20190820eq2}) can be rewritten as 
\begin{equation}
\emph{\textbf{b}}=\textbf{A}\emph{\textbf{d}} 
\label{20190820eq3}
\end{equation}
where 
\begin{equation}
\begin{aligned}
\textbf{A}&\triangleq \textbf{P}\\
\emph{\textbf{b}}&\triangleq \Bigg[p\left(x_1,y_1\right)-\emph{\textbf{r}}_1^T\boldsymbol{\Omega}, q\left(x_1,y_1\right)-\emph{\textbf{s}}_1^T\boldsymbol{\Omega},..., \\
&p\left(x_N,y_N\right)-\emph{\textbf{r}}_N^T\boldsymbol{\Omega}, q\left(x_N,y_N\right)-\emph{\textbf{s}}_N^T\boldsymbol{\Omega}\Bigg],
\end{aligned}\nonumber
\end{equation}
where the meanings of $\emph{\textbf{r}}_i$ and $\emph{\textbf{s}}_i$ are listed in Table \ref{20190820table1}.
The least square solution of (\ref{20190820eq3}) is 
\begin{equation}
\hat{\emph{\textbf{d}}}=\left(\textbf{A}^T\textbf{A}\right)^{-1}\textbf{A}^T\emph{\textbf{b}} 
\label{20190822eq1}
\end{equation}
To simplify the description of the bias expression, let
\begin{equation}
\begin{aligned}
\textbf{M}\triangleq \textbf{A}^T\textbf{A} &= diag\left[\left(x_i-x_f\right)^2+\left(y_i-y_f\right)^2\right]_{N\times N}\\
&= diag\left[m_{ii}\right]_{i=1...N}
\end{aligned}\nonumber
\end{equation}
\begin{equation}
\begin{aligned}
\emph{\textbf{V}}\triangleq \textbf{A}^T\emph{\textbf{b}} &= diag\left[\left(x_i-x_f\right)v_{pi} +\left(y_i-y_f\right)v_{qi}\right]^T\\
&\triangleq \left[v_1...v_N\right]^T
\end{aligned}\nonumber
\end{equation}
where $v_{pi}= p\left(x_i,y_i\right)-\emph{\textbf{r}}_i^T\boldsymbol{\Omega}$, $v_{qi}= q\left(x_i,y_i\right)-\emph{\textbf{s}}_i^T\boldsymbol{\Omega}$ and $diag []$ represents the diagonal matrix.

\begin{figure} \centering 
\includegraphics[scale=0.27]{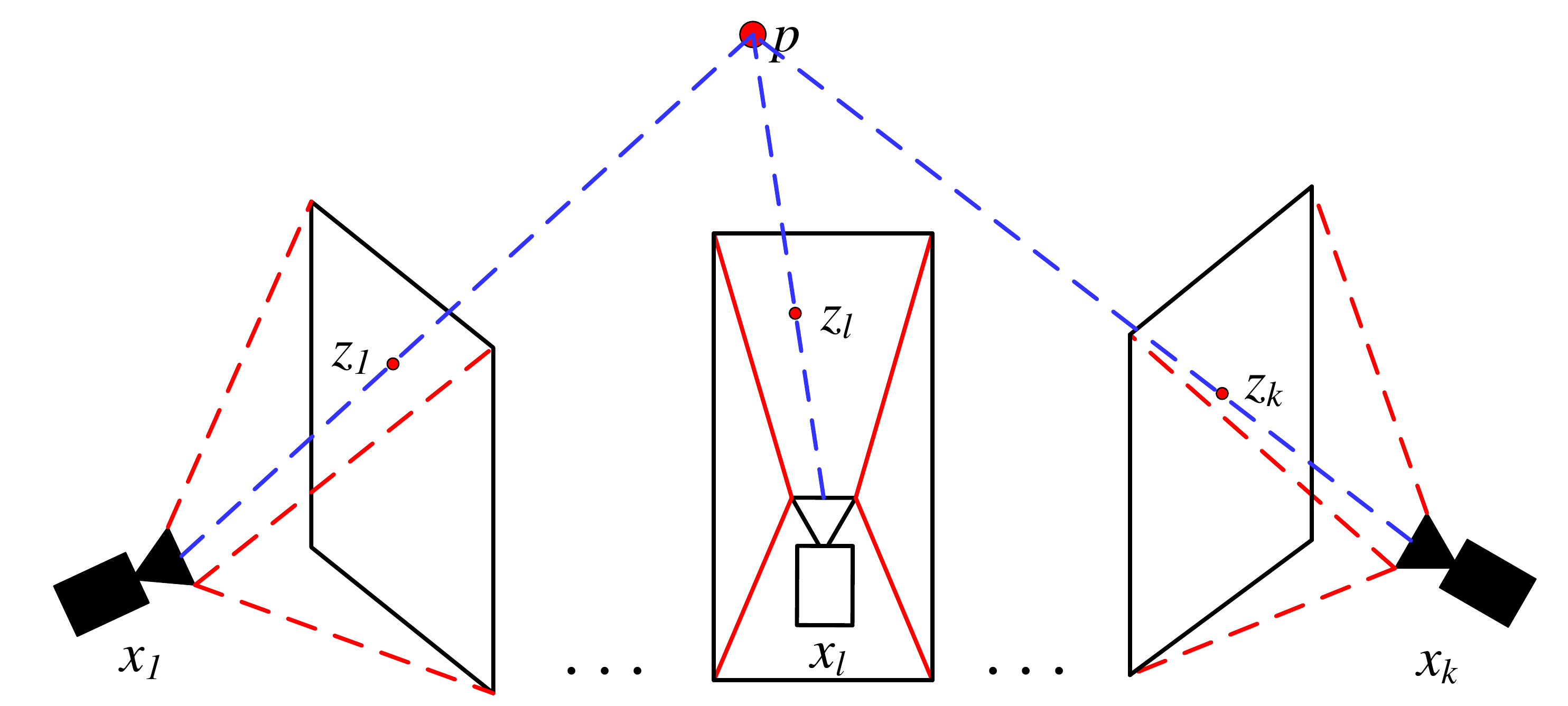} 
\caption{Recover map points using $k$ frames in VSLAM. In this figure, the map point $\emph{\textbf{p}}$ is projected to the image point $\emph{\textbf{z}}$. If $\emph{\textbf{p}}$ can be observed by $k$ images, then it can be jointly recovered by these $k$ images.}
\label{20190821fig2} 
\end{figure}

According to \cite{41}, the bias of the inverse depth estimation of the $i^{th}$ feature point is 
\begin{equation}
\begin{aligned}
\mu\left(\hat{\emph{\textbf{d}}}_i\right)&=\frac{2\sigma_i^2v_i}{m_{ii}^2}\\
&+\frac{2\sigma_i^2}{m_{ii}^2}\left[\left(x_i-x_f\right)^2\emph{\textbf{r}}_{ix}^T+ \left(y_i-y_f\right)^2\emph{\textbf{s}}_{iy}^T \right]\boldsymbol{\Omega}\\
&+\frac{2\sigma_i^2}{m_{ii}^2}\left[\left(x_i-x_f\right)\left(y_i-y_f\right)\left(\emph{\textbf{r}}_{iy}^T+\emph{\textbf{s}}_{ix}^T \right)\right]\boldsymbol{\Omega}\\
&+\frac{\sigma_i^2}{m_{ii}^2}\left[\left(x_i-x_f\right)\omega_y - \left(y_i-y_f\right)\omega_x-\left(\emph{\textbf{r}}_{ix}^T+\emph{\textbf{s}}_{iy}^T \right)\boldsymbol\Omega\right]
\label{20190822eq2}
\end{aligned}
\end{equation}
where $\sigma_i^2$ is the variance in the image coordinate measurements, $\emph{\textbf{r}}_{ix}$ and $\emph{\textbf{s}}_{ix}$ is the derivative of $\emph{\textbf{r}}_{i}$ and $\emph{\textbf{s}}_{i}$ with respect to $x$, and $\emph{\textbf{r}}_{iy}$ and $\emph{\textbf{s}}_{iy}$ have similar meanings.

Since the depth result is easily affected by the noise for two-frame reconstruction, $L$ two-frame reconstruction results are employed to reduce the depth error, where the depth result and the bias are
\begin{equation}
\hat{\emph{\textbf{d}}}=\frac{1}{L}\sum_{j=1}^{L}\hat{\emph{\textbf{d}}}^j
\label{20190820eq4}
\end{equation}
\begin{equation}
\boldsymbol{\mu}\left(\hat{\emph{\textbf{d}}}\right)=\frac{1}{L}\sum_{j=1}^{L}\boldsymbol{\mu}\left(\hat{\emph{\textbf{d}}}^j\right)
\label{20190820eq5}
\end{equation}
For the biased estimation, $E\left[\hat{\emph{\textbf{d}}}\right]=\overline{\emph{\textbf{d}}} + \boldsymbol{\mu}\left(\hat{\emph{\textbf{d}}}\right)$, where $\overline{\emph{\textbf{d}}}$ is the truth-value. If $\hat{\emph{\textbf{d}}}_c=\hat{\emph{\textbf{d}}} - \boldsymbol{\mu}\left(\hat{\emph{\textbf{d}}}\right)$, then $E\left[\hat{\emph{\textbf{d}}}_c\right]=E\left[\hat{\emph{\textbf{d}}}\right] - \boldsymbol{\mu}\left(\hat{\emph{\textbf{d}}}\right) = \overline{\emph{\textbf{d}}}$. $\hat{\emph{\textbf{d}}}_c$ is an unbiased estimation.

In VSLAM, map points can be observed by multi-frames, which is shown in Fig. \ref{20190821fig2}. Therefore, map points are jointly recovered by these frames, and $\boldsymbol{\mu}\left(\hat{\emph{\textbf{d}}}\right)$ may not be the bias value of map points due to it is derived based on two-frame. To solve this problem, we first reconstruct map points using multi-two-frame, and getting $\hat{\emph{\textbf{d}}}$ and $\boldsymbol{\mu}\left(\hat{\emph{\textbf{h}}}\right)$ according to (\ref{20190820eq4}) and (\ref{20190820eq5}). Then $\hat{\emph{\textbf{d}}}_c$ can be obtained. Since $\hat{\emph{\textbf{d}}}_c$ is an unbiased estimation, suppose the actual value acquired by multi-frame is $\widetilde{\emph{\textbf{d}}}$, the bias value of map points is $\boldsymbol{\mu}\left(\widetilde{\emph{\textbf{d}}}\right) = \widetilde{\emph{\textbf{d}}} - \hat{\emph{\textbf{d}}}_c$. According to the bias value $\boldsymbol{\mu}\left(\widetilde{\emph{\textbf{d}}}\right)$ and (\ref{eq6}), the camera pose estimation error can be corrected.

\begin{figure}[htb]
 \centering 
\includegraphics[scale=0.3]{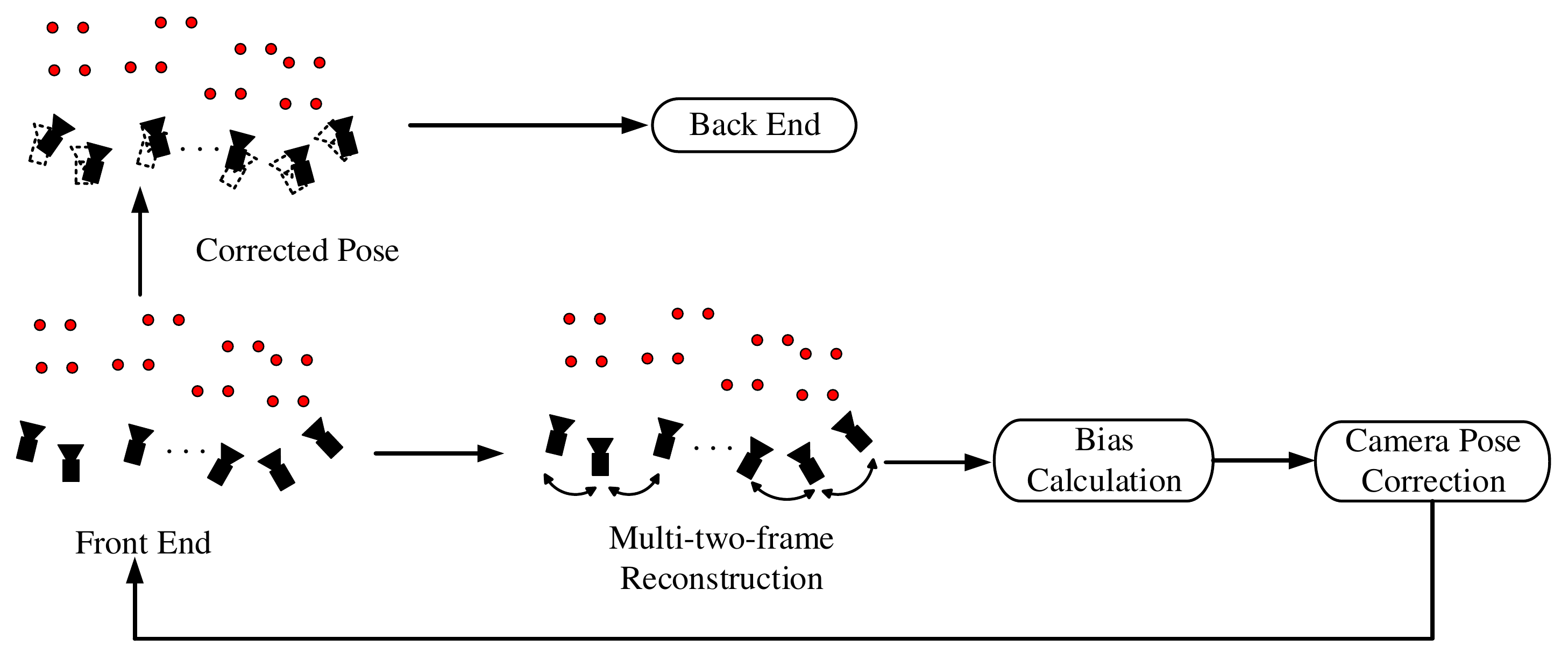} 
\caption{A block diagram illustrating the full pipeline of our camera pose correction method.}
\label{20190822fig1} 
\end{figure}

\begin{table*}[htb]
\newcommand{\tabincell}[2]{\begin{tabular}{@{}#1@{}}#2\end{tabular}}
\center
\caption{The RMSE error of camera pose estimation results (unit: m).}\label{table2}
{
\setlength{\tabcolsep}{5mm}{\begin{tabular}{ccccccc}
\hline  
Sequence & \tabincell{c}{Our method\\ no loop} & \tabincell{c}{VINS-Mono\\ no loop}& \tabincell{c}{VINS-GF \\no loop} & \tabincell{c}{Our method \\with loop} & \tabincell{c}{VINS-Mono \\with loop}  & \tabincell{c}{VINS-GF \\with loop}\\
\hline  
MH\_01\_easy	&0.20&	0.15& 0.24&	\textbf{0.06}&	0.12  & 0.22\\
MH\_02\_easy	    &\textbf{0.15}&	0.15& 0.15&	\textbf{0.07}&	0.12   & 0.47\\
MH\_03\_medium	&\textbf{0.21}&	0.22& 0.27&	\textbf{0.078}&	0.13  & 0.17\\
MH\_04\_difficult	&\textbf{0.31}&	0.32& 0.36 &	\textbf{0.11}&	0.18 & 0.16\\
MH\_05\_difficult	&\textbf{0.25}&	0.30&0.29 &	\textbf{0.12}&	0.21 &0.17\\
V1\_01\_easy	    &0.089&	0.079& 0.095&	\textbf{0.048}&	0.068  & 0.070\\
V1\_02\_medium	&\textbf{0.09}&	0.11& 0.096&	\textbf{0.050}&	0.084  & 0.068\\
V1\_03\_difficult	&\textbf{0.15}&	0.18& 0.20&	0.212&	0.19  & 0.195\\
V2\_01\_easy	    &0.092&	0.080 & 0.093 &	\textbf{0.050}&	0.081& 0.081\\
V2\_02\_medium	&\textbf{0.13}&	0.16& 0.13&	\textbf{0.088}&	0.16  & 0.18\\
\hline 
Average error & \textbf{0.1671} & 0.1799 & 0.1924 & \textbf{0.0886} & 0.1343 & 0.1784\\
\hline 
\end{tabular}}}\end{table*}

\begin{table*}[htb]
\newcommand{\tabincell}[2]{\begin{tabular}{@{}#1@{}}#2\end{tabular}}
\center
\caption{The median error of camera pose estimation results (unit: m).}\label{table3}
\setlength{\tabcolsep}{5mm}{\begin{tabular}{ccccccc}
\hline  
Sequence & \tabincell{c}{Our method\\ no loop} & \tabincell{c}{VINS-Mono\\ no loop} & \tabincell{c}{VINS-GF \\no loop} & \tabincell{c}{Our method \\with loop} & \tabincell{c}{VINS-Mono \\with loop}& \tabincell{c}{VINS-GF \\with loop}\\
\hline  
MH\_01\_easy&	\textbf{0.16292}&	0.17504& 0.19594 &	\textbf{0.05087}&	0.12 & 0.15811\\
MH\_02\_easy&	0.11173& 0.08014& 0.08317 & 0.05377& 0.06140 & 0.34695\\			
MH\_03\_medium&	\textbf{0.18585}&	0.20008& 0.20036&	\textbf{0.05888}&	0.07544   & 0.11668\\
MH\_04\_difficult&	\textbf{0.35914}&	0.40284 & 0.38965&	\textbf{0.10408}&	0.11880 & 0.13352\\
MH\_05\_difficult	&0.25327&	0.23488& 0.29118&	\textbf{0.10486}	&0.17543  & 0.14442\\
V1\_01\_easy&	\textbf{0.07016}&	0.07120& 0.07257&	\textbf{0.04345}&	0.05756  & 0.06305\\
V1\_02\_medium&	\textbf{0.08712}&	0.09171& 0.09319 &	\textbf{0.04549}&	0.05951 & 0.06257\\
V1\_03\_difficult&	\textbf{0.11329}&	0.16920& 0.18705 &	0.16260&	0.16229 & 0.17792\\
V2\_01\_easy&				0.06844& 0.05234 & 0.06680& 0.04313& 0.05099 & 0.06538\\
V2\_02\_medium&	\textbf{0.08515}&	0.08650& 0.08644&	\textbf{0.07149}&	0.11029  & 0.12626\\
\hline 
Average error & \textbf{0.14970} & 0.15639 & 0.16663 & \textbf{0.06337} & 0.09917 & 0.13948\\
\hline 
\end{tabular}}
\end{table*}

\subsection{Anomaly detection strategy}
In practical applications, bias values of some map points may be very large due to mismatching or camera motion error, which can make the system unstable. To avoid this negative effect, we propose a heuristic strategy to determine which map points can be used to compensate the pose estimation error. Since the bias value is negative correlation to the image parallel, the principle of our strategy is choosing map points recovered by large parallels. In this article, our method is combined with the VINS-Mono \cite{00} where the camera motion is detected by the IMU. Therefore, the parallel can be replaced by the camera angular velocity. The strategy is 
\begin{equation}
thresh\ of\ bias =\left\{
\begin{array}{rcl}
0.1m      &      & {if \  \lVert \boldsymbol{\omega}_{cam} \rVert_2 \geq 0.5 }\\
0.3m   &      & {if \  0.3 \leq \lVert \boldsymbol{\omega}_{cam} \rVert_2 < 0.5}\\
0.5m     &      & {if \   \lVert \boldsymbol{\omega}_{cam} \rVert_2 < 0.3}
\end{array} \right.
\end{equation}
where $\lVert \boldsymbol{\omega}_{cam} \rVert_2$ represents the $l_2$-norm of the angular velocity. If the bias value of the map point is larger than the thresh value, the corresponding map point is abandoned to correct the pose error.
These thresh values are set based on testing results in different datasets, which makes our algorithm get stable and accurate results.

The full pipeline of our camera pose correction method is shown in Fig. \ref{20190822fig1}. In this figure, the camera pose obtained by the front end is used to calculate the depth and bias of map points according to (\ref{20190822eq1}) and (\ref{20190822eq2}). To make full use of all frames which observes the same map points, multi-two-frame reconstructions are processed. Based on reconstruction results, bias value is calculated by (\ref{20190820eq4}) and (\ref{20190820eq5}), and the camera pose error in front end is corrected. Finally the results of the front end are optimized by the back end. Our algorithm only modifies the results obtained by the front end and is independents of the back end. Therefore, it is easy to integrate into different VO/SLAM systems.

\textbf{Remark 1}: In abstract and Section \uppercase\expandafter{\romannumeral1}, we emphasis that one of characteristics of our method is all original system information is reserved. It is not contradict some map points may be abandoned to correct the pose error, because pose correction is a separate procedure. This means that steps of the original system are not changed, therefore, all system information still can be used after the pose correction.

\begin{figure*}[htb]
 \centering 
\subfigure[MH\_01\_easy] { \label{fig3} 
\includegraphics[scale=0.18]{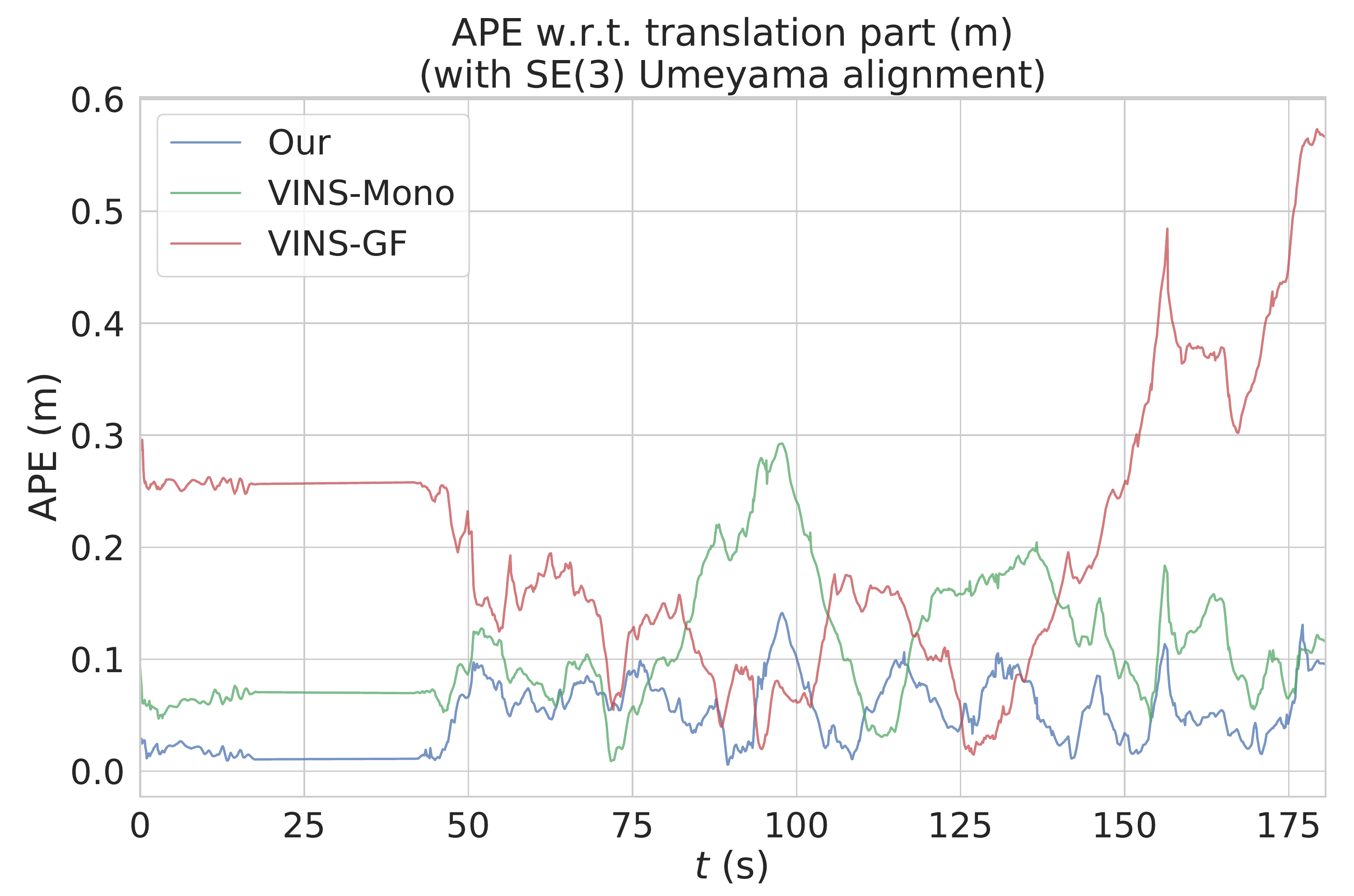} 
} 
\subfigure[MH\_05\_difficult] { \label{fig3} 
\includegraphics[scale=0.18]{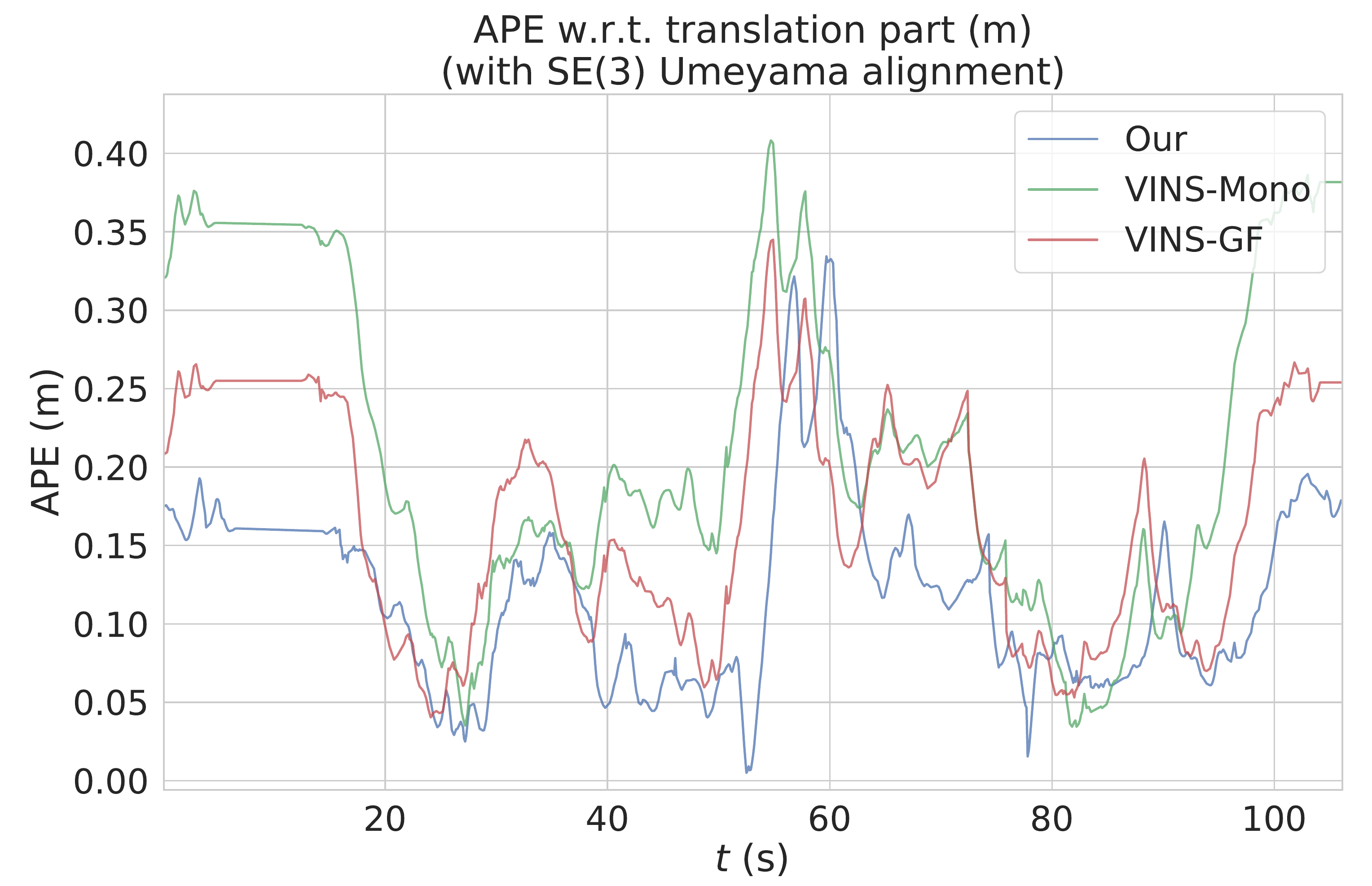} 
} 
\subfigure[V2\_02\_medium] { \label{fig3} 
\includegraphics[scale=0.18]{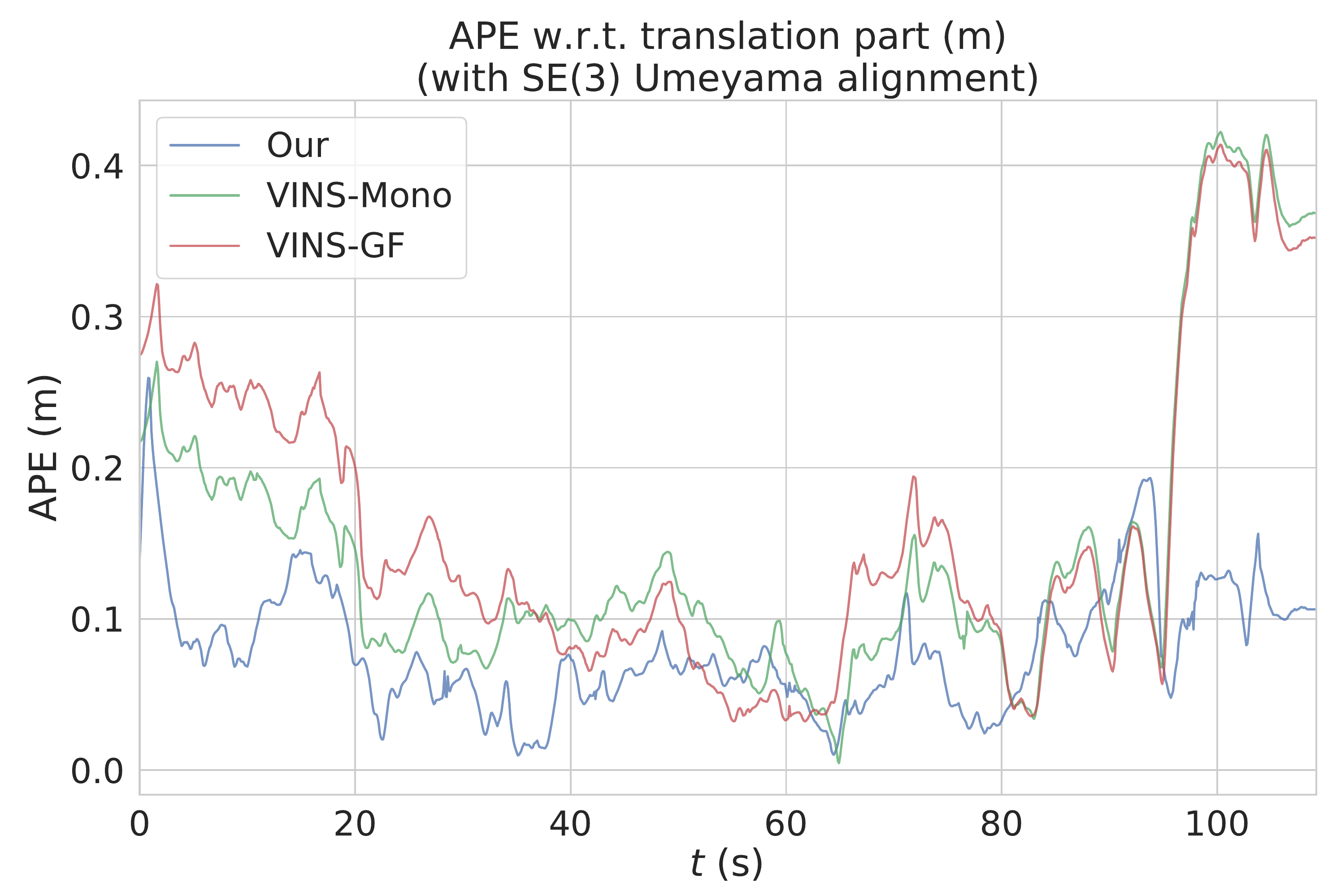} 
} 
\caption{The curves of absolute pose error (APE) w.r.t. translation part of sequences MH\_01\_easy, MH\_05\_difficult and V2\_02\_medium. The results are acquired in the system with loop detection.}
\label{fig4} 
\end{figure*}
\begin{figure*}[htb]
 \centering 
\subfigure[MH\_01\_easy] { \label{fig3} 
\includegraphics[scale=0.15]{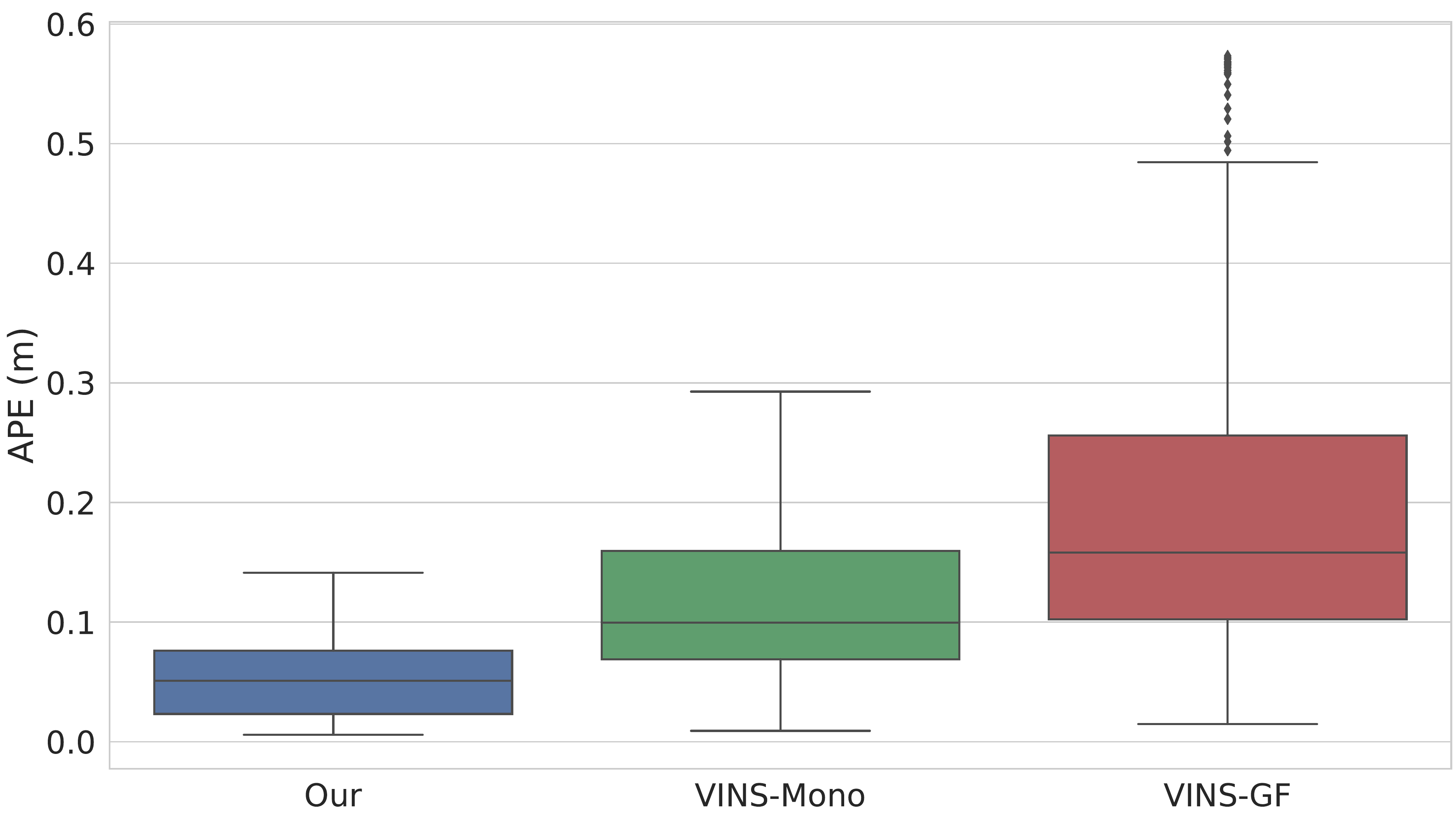} 
} 
\subfigure[MH\_05\_difficult] { \label{fig3} 
\includegraphics[scale=0.15]{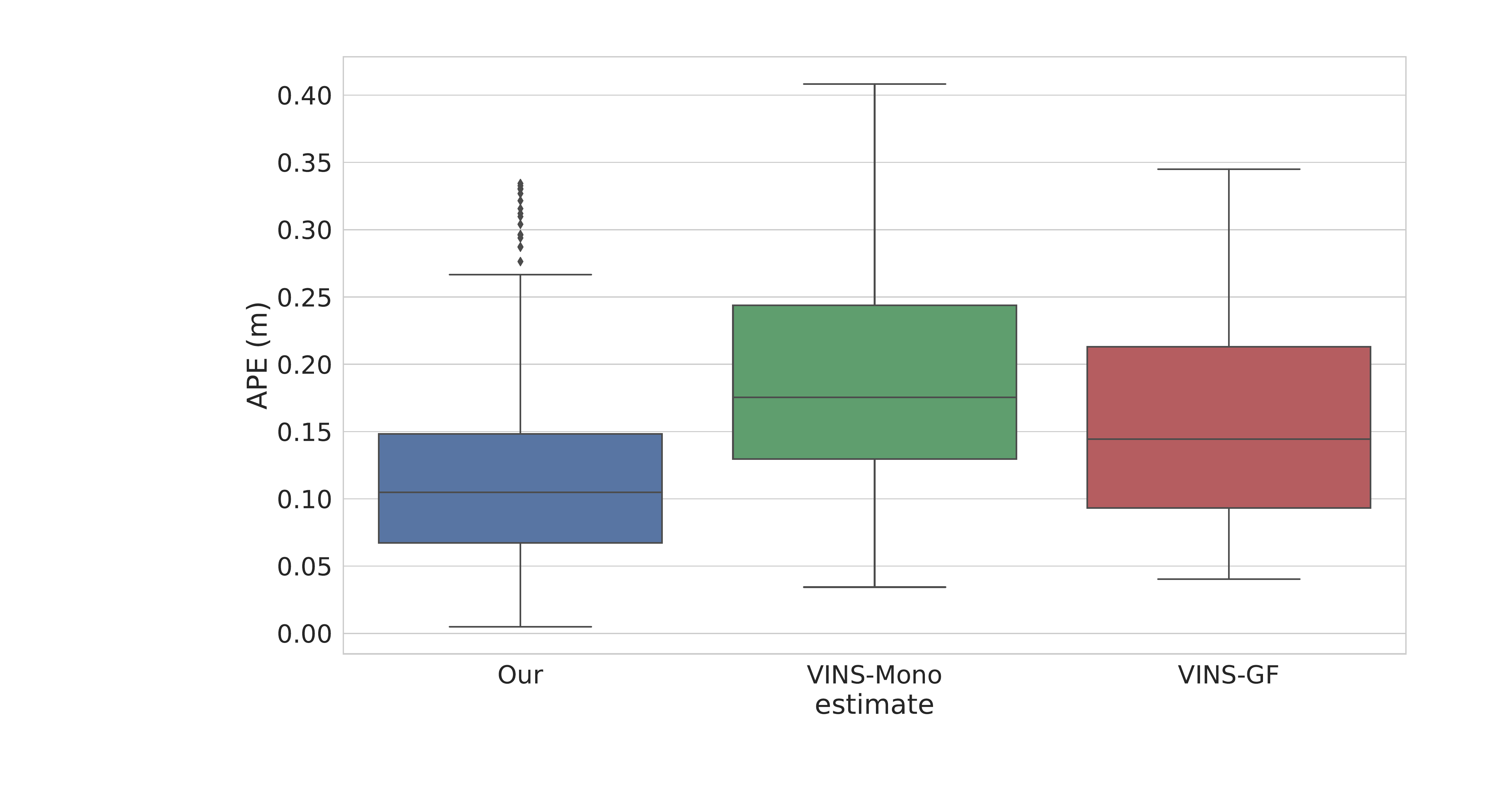} 
} 
\subfigure[V2\_02\_medium] { \label{fig3} 
\includegraphics[scale=0.15]{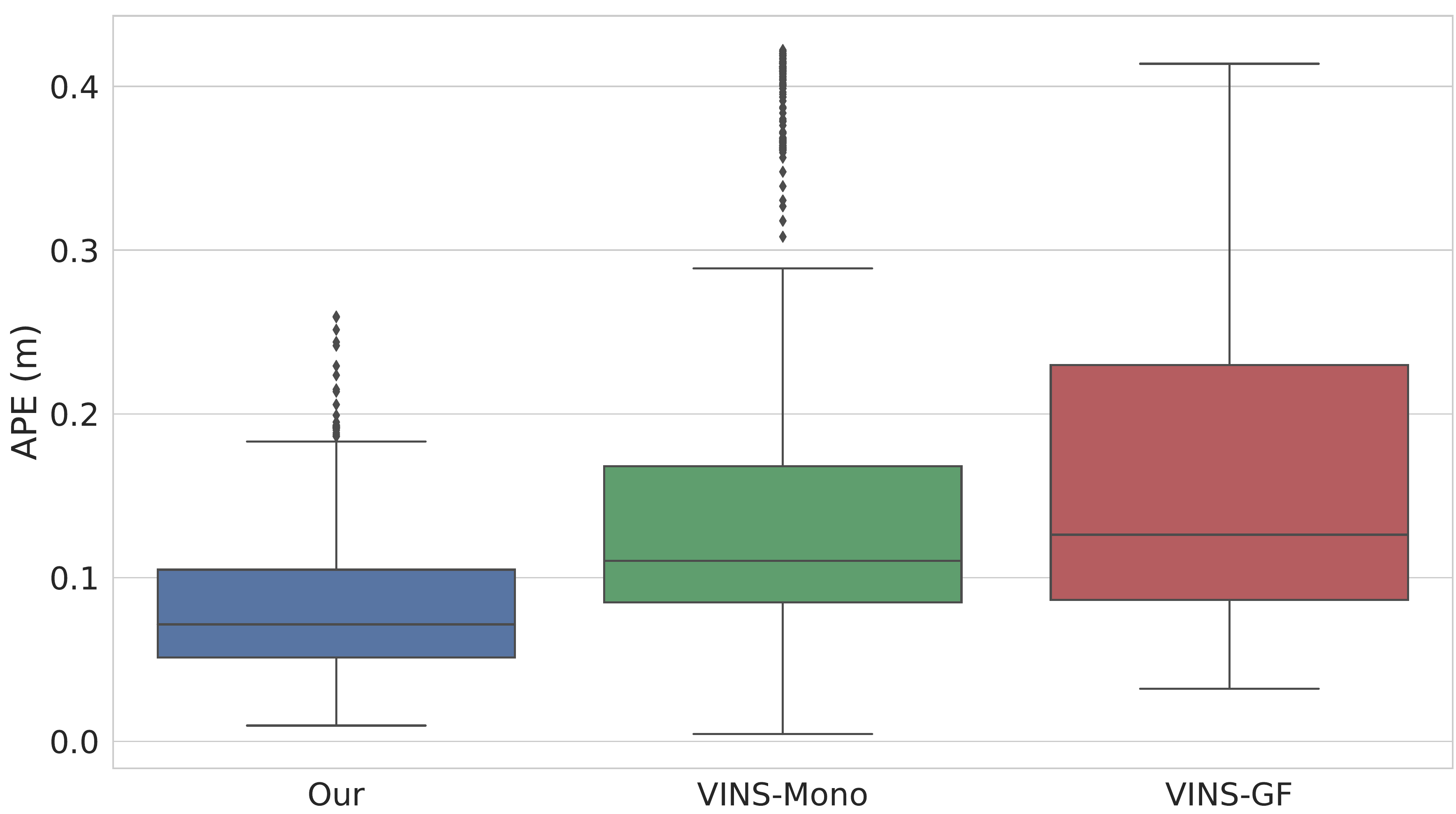} 
} 
\caption{The box plot of absolute pose error (APE) w.r.t. translation part corresponding to Fig. \ref{fig4}.}
\label{fig5} 
\end{figure*}
\begin{figure}[htb]
 \centering 
\includegraphics[scale=0.2]{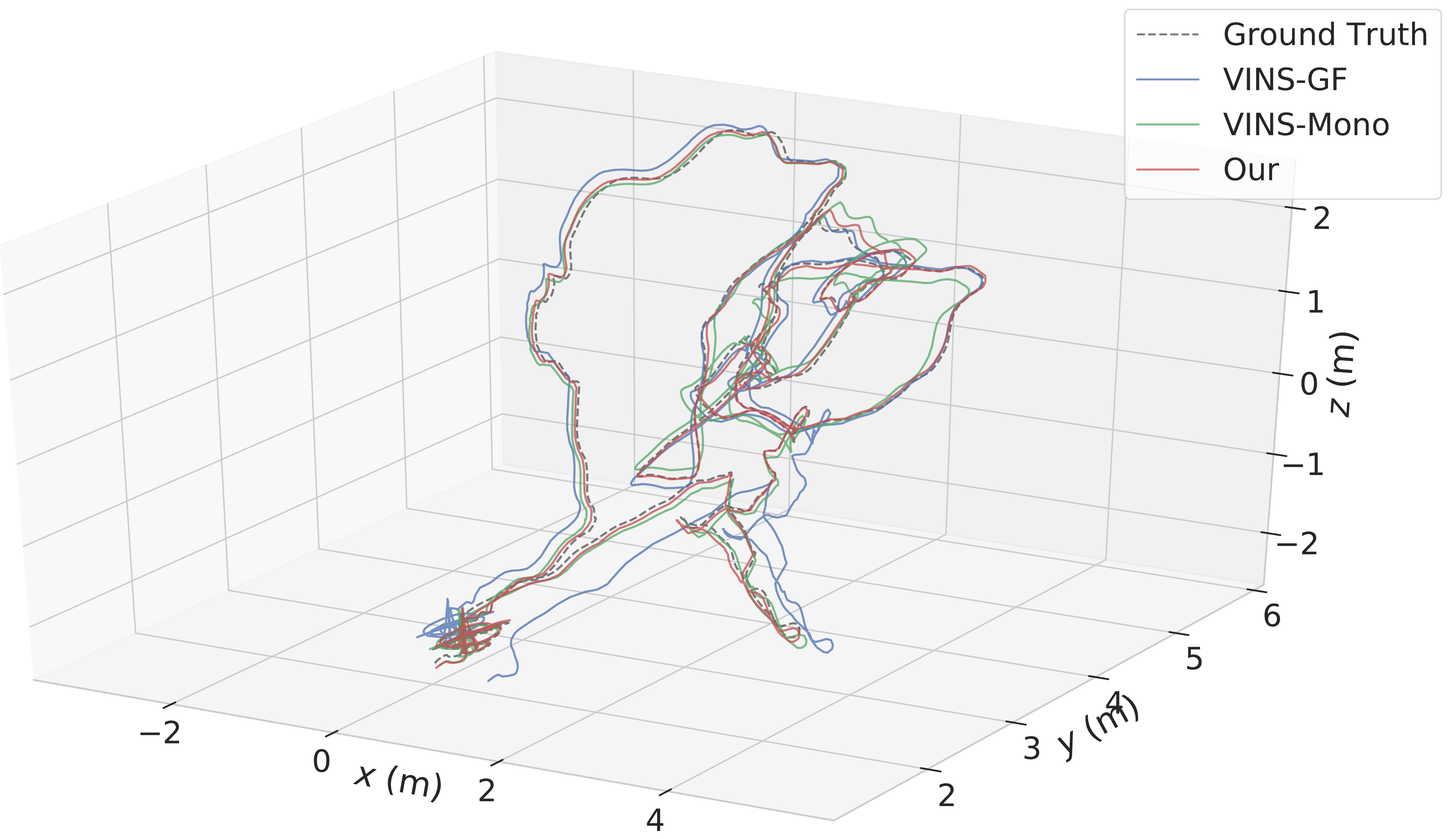} 
\caption{The trajectory of sequence MH\_01\_easy.}
\label{fig6} 
\end{figure}

\section{Experiments}

The effectiveness of our method is verified by integrating it into the VINS-Mono framework. Our algorithm is activated after finishing initialization. The pose error correction is implemented if feature points are successfully tracked and triangulated, and finally BA optimization is executed. 
The experiments are performed in an Intel Core i7-26700QM computer with 8GB RAM, and we evaluate the accuracy in the EuroC dataset \cite{33}. 
To make results more reliable, we run five times in each sequence for every compared method and calculate the mean value as their final results. 

\subsection{Compared results with VINS-Mono based methods}
The accuracy of our method is compared with the original VINS-Mono and the good feature selection method (GF) \cite{31} mentioned at the end of Section \uppercase\expandafter{\romannumeral1}. Based on the published source code of \cite{31}, which is realized based on the ORB-SLAM, we combine it with the VINS-Mono and set the number of selected map points in the sliding window to 50. This value is large enough to make the estimation result reach a good balance between the accuracy and confidence. If the number of map points in the sliding window is less than 50, all of them are used in BA optimization to avoid the side effect caused by too little map points. Other parameters in these methods are the same. The translation Root Mean Square Error (RMSE) and the median errors of the keyframe trajectory for each sequence are shown in Table \ref{table2} and Table \ref{table3}, respectively. The first three columns represent the results without loop detection, and the results with loop detection are given in the last three columns.

It can be seen that our method performs best on at least seven of ten sequences no matter in the system with or without loop detection. Especially for the RMSE result with loop detection, the accuracies in nine sequences are largely improved, e.g. 0.12$m$ to 0.06$m$ in the sequence MH\_01\_easy and 0.16$m$ to 0.088$m$ in the sequence V2\_02\_medium. The difference to the best system on the other sequences is small. For the GF method, due to only a subset of map points is employed in BA optimization, the robustness of the estimation result to the IMU noise is reduced. That is the reason why the performance of VINS-Mono decreases after combined with the GF method. The compared results with the GF algorithm demonstrate the advantage of our method, i.e. robustness to system noise. It guarantees our system can still achieve a good performance in systems contain the IMU or little map points. 

To show the details of pose estimation error, we visualize the absolute pose error (APE) w.r.t. translation part of sequences MH\_01\_easy, MH\_05\_difficult and V2\_02\_medium in Fig. \ref{fig4} and Fig. \ref{fig5}.
According to Fig. \ref{fig4}, it is obvious that our method has the smallest translation error, especially at the beginning and the end of the trajectory. This result shows superiority of our method in suppressing the accumulative error. Fig. \ref{fig5} shows the box plot of the translation error. In this figure, the dark line represents the mean value of the error, and the height of the box represents the variance. From this figure, the mean value of our method is less than 0.1$m$ in sequences MH\_01\_easy and V2\_02\_medium and about 0.1$m$ in the sequence MH\_05\_difficult, while the mean value of compared methods are more than 0.1$m$ in sequences MH\_01\_easy and V2\_02\_medium and more than 0.15$m$ in the sequence MH\_05\_difficult. Meanwhile, the variance of our method is also the smallest. This result further demonstrates the effectiveness and stability of our method in reducing the pose estimation error. 

We also show the trajectory of the sequence MH\_01\_easy with loop detection in Fig. \ref{fig6}. In this figure, due to the influence of IMU noise, the GF method cannot return to the original point at the end of the trajectory.

\subsection{Compared results with other VIO methods}
\begin{table}[htb]
\center
\caption{The RMSE error of camera pose estimation results (unit: m).}\label{table4}
\setlength{\tabcolsep}{1mm}{\begin{tabular}{ccccccc}
\hline  
Sequence & Our method & R-VIO & ROVIO & MSCKF& OKVIS\\
\hline  
MH\_01\_easy	&0.20&	0.38& 0.21 & 0.42& 0.16\\
MH\_02\_easy	    &\textbf{0.15}&	0.74& 0.25 & 0.45&0.22\\
MH\_03\_medium	&\textbf{0.21}&	0.35& 0.25 & 0.23&0.24\\
MH\_04\_difficult	&\textbf{0.31}&	1.03& 0.49 & 0.37&0.34\\
MH\_05\_difficult	&\textbf{0.25}&	0.85&0.52 &0.48&0.47\\
V1\_01\_easy	    &0.089&	0.085& 0.10&0.34&0.09\\
V1\_02\_medium	&\textbf{0.09}&	0.15& 0.10&0.20&0.20\\
V1\_03\_difficult	&0.15&	0.13& 0.14&0.67&0.24\\
V2\_01\_easy	    &0.092&	0.080 & 0.12&0.10&0.13\\
V2\_02\_medium	&\textbf{0.13}&	0.16& 0.14&0.16&0.16\\
\hline 
Average error & \textbf{0.1671} & 0.3955 & 0.232&0.342 &0.225\\
\hline 
\end{tabular}}
\end{table}
The accuracy of our method is also compared with another four state-of-the-art visual inertial odometry (VIO) algorithms, R-VIO \cite{34}, ROVIO \cite{35}, MSCKF \cite{39} and OKVIS \cite{40}. Since loop detection is not available in some of them, only RMSE results without loop detection are compared. The result of R-VIO comes from the original literature, and \cite{36} presents the results of the other methods tested on the laptop. Their hardware platforms are better than ours, and compared results are listed in Table \ref{table4}. 

In this table, our method performs best on six of ten sequences compared with R-VIO and performs best on at least nine of ten sequences compared with the other methods. The difference to the best system on the other sequences is small. Our method also has the smallest average error (0.1671$m$ for our, 0.3955$m$ for R-VIO, 0.232$m$ for ROVIO, 0.342$m$ for MSCKF and 0.225$m$ for OKVIS).

\subsection{Time-consuming results}
\begin{table}[htb]
\center
\caption{The average time-consuming of solveOdometry function for solving one keyframe (unit: millisecond).}\label{table5}
\setlength{\tabcolsep}{2.2mm}{\begin{tabular}{cccc}
\hline  
Sequence & Our method & VINS-Mono & Time difference\\
\hline  
MH\_01\_easy	&61.51&	49.32& 12.19\\
MH\_02\_easy	    &61.39&	48.44& 12.95\\
MH\_03\_medium	&59.56&	48.59& 10.97\\
MH\_04\_difficult	&57.32&	48.72& 9.12\\
MH\_05\_difficult	&59.93&	47.25&12.68\\
V1\_01\_easy	    &61.35&	50.01& 11.34\\
V1\_02\_medium	&48.38&	44.28& 4.10\\
V1\_03\_difficult	&40.74&	37.09& 3.65\\
V2\_01\_easy	    &56.78&	48.03 & 8.75\\
V2\_02\_medium	&55.03&	42.22& 12.81\\
\hline 
Average time &  56.20 & 46.39 & 9.81\\
\hline
\end{tabular}}
\end{table}
The time-consuming of our method is compared with the original VINS-Mono. Our method is integrated into the solveOdometry function in vins\_estimator.cpp file, and other files are not modified. Therefore, it is enough to test the time difference of this function, which is more convenient and accurate than testing the whole system. We run five times in each sequence for every method and calculate the mean value as the final results. The average time-consuming of solveOdometry function to solve one keyframe is shown in Table \ref{table5}.

In this table, the time differences of sequences V1\_02\_medium, V1\_03\_difficult and V2\_01\_easy are smaller than the other sequences. Since bias calculating occupies most time of our algorithm, different map point number leads to different time-consuming. The texture of Vicon Room is simple than Machine Hall, and the camera rotation is faster than the other sequences in these small time difference sequences. Therefore, less map points are recovered, and the added time to solveOdometry function is little.

According to Table \ref{table5}, it can be seen that the average time-consuming of our method is larger 9.81$ms$ than the VINS-Mono. This value is acceptable for practical applications.

\section{conclusion}
This paper proposes a camera pose correction method, which is compact and effective and reserves all system information. The relationship between the pose estimation error and the biased value of the map point is derived. Based on this relationship and the bias calculating method, the pose estimation results are corrected. We verify the effectiveness and efficiency of our method by comparing with other state-of-the-art algorithms. The future work will integrate our method into other visual SLAM systems, which does not contain IMU information.

\end{document}